\documentclass[times,10pt,twocolumn,notitlepage]{article}

\usepackage{graphicx}
\usepackage{amssymb}
\usepackage{amsmath}
\usepackage{latexsym}

\usepackage{url}
\usepackage{xcolor}
\definecolor{newcolor}{rgb}{.8,.349,.1}

\graphicspath{{./figures_for_arXiv/}}

\title{Fine Hand Segmentation using Convolutional Neural Networks}

\author{
  Tadej Vodopivec$^{1,2}$ \quad Vincent Lepetit$^1$ \quad Peter Peer$^2$\\
  {\small  $^1$ Institute for Computer Graphics and Vision, Graz University of
    Technology, Austria}\\
  {\small $^2$ Faculty of Computer and Information Science,
    University of Ljubljana, Ve\v{c}na pot 113, 1000 Ljubljana, Slovenia}\\
}
\date{}

\begin{document}

\maketitle

\section*{Abstract}
We  propose  a  method  for  extracting very  accurate  masks  of  hands  in
egocentric  views.    Our  method  is   based  on  a  novel   Deep  Learning
architecture: In contrast with current Deep  Learning methods, we do not use
upscaling layers  applied to a  low-dimensional representation of  the input
image. Instead, we  extract features with convolutional layers  and map them
directly to a  segmentation mask with a fully connected  layer. We show that
this approach, when  applied in a multi-scale fashion, is  both accurate and
efficient enough for real-time.  We demonstrate  it on a new dataset made of
images captured in various environments, from the outdoors to offices.

\section{Introduction}

To ensure that the user perceives the  virtual objects as part of the real world
in  Augmented   Reality  applications,  these   objects  have  to   be  inserted
convincingly enough. By far, most of the research in this direction has focus on
3D pose estimation, so that the object  can be rendered at the right location in
the  user's view~\cite{Vacchetti04a,Klein07,Newcombe11b}.   Other  works aim  at
rendering the light  interaction between the virtual objects and  the real world
consistently~\cite{Debevec98,Meilland13}.

Significantly less  works have  tackled the problem  of correctly  rendering the
occlusions which occur when a real object  is located in front of a virtual one.
\cite{Lepetit00} provides  a method that  requires interaction with a  human and
works only for  rigid objects.  \cite{Pilet07} relies  on background subtraction
but this  is prone to fail  when foreground and background  have similar colors.
Depth cameras bring  now an easy solution to handling  occlusions, however, they
provide a poorly  accurate 3D reconstruction of the occluding  boundaries of the
real  objects,  which are  essential  for  a  convincing perception.  The  human
perception  is actually  very  sensitive  to small  deviations  from the  actual
locations    in     occlusion    rendering,    making    the     problem    very
challenging~\cite{Zollmann10}.

With the development of hardware such as the HoloLens, which provides precise 3D
registration and  crisp rendering of  the virtual objects,  egocentric Augmented
Reality  applications  can be  foreseen  to  become  very  popular in  the  near
future. This  is why  we focus here  on correct rendering  of occlusions  by the
user's hands of the virtual objects. More  exactly, we assume that the hands are
always  in  front   of  the  virtual  objects,  which  is   realistic  for  many
applications, and  we aim at  estimating a pixel-accurate  mask of the  hands in
real-time.

The last years have seen the development of different segmentation methods based
on  Convolutional Neural  Networks~\cite{Long15,Badrinarayanan15}.  While  our
method also  relies on Deep  Learning, its architecture has  several fundamental
differences.    It   is   partially  inspired   by   Auto-Context~\cite{Tu09}:
Auto-Context is a segmentation method in  which a segmenter is iterated, and the
segmentation  result of  the previous  step  is used  in the  next iteration  in
addition to the original image.

The fundamental difference between our approach and the original Auto-Context is
that the initial segmentation is performed  on a downscaled version of the input
image. The  resulting segmentation is then  upscaled before being passed  to the
second  iteration.  This  allows  us  to take  the  context  into  account  very
efficiently. We can also obtain precise localization of the segmentation
boundaries, because we avoid using pooling.

In the remainder of the paper, we  first discuss related work, then describe our
method, and finally  present and discuss our  results on a new  dataset for hand
segmentation.


\section{Related Work}

Hand segmentation is a  very challenging task as hands can  be very different in
shape and skin color, look very different under another viewpoint, can be closed
or open, can be partially occluded, can have different positions of the fingers,
can be grasping objects or other hands, etc.

Skin  color   is  a  very  obvious   cue~\cite{skin_color_one,  skin_color_two},
unfortunately, this approach is prone to fail  as other objects may have a similar
color. Other approaches  assume that the camera is static  and segment the hands
based on their movement~\cite{hand_movement}, use  a simple or even single-color
background~\cite{hand_simple_background}, or rely  on depth information obtained
by  an RGB-D  camera~\cite{hand_depth}.  None  of these  approaches can  provide
accurate masks in general conditions.

The  method we  propose is  based on  convolutional neural  networks~\cite{cnn}.
Deep Learning has already been applied to segmentation, and recent architectures
tend to be made  of two parts: The first part  applies convolutional and pooling
layers to the  input image to produce a  compact, low-resolution representation;
the second part applies deconvolutional layers to this representation to produce
the  final  segmentation, at  the  same  resolution  as  the input  image.  This
typically results in oversmoothed segments, which we avoid with our approach.

\section{Method}

In  this section,  we  describe  our approach.   We  first  present our  initial
architecture  based on  multiscale analysis  of the  input. We  then split  this
architecture in two to obtain our more efficient, final architecture. We finally
detail our methodology to select the meta-parameters of this architecture.

\subsection{Initial Network Architecture}

\begin{figure}
  \begin{center}
    \includegraphics[width=\columnwidth]{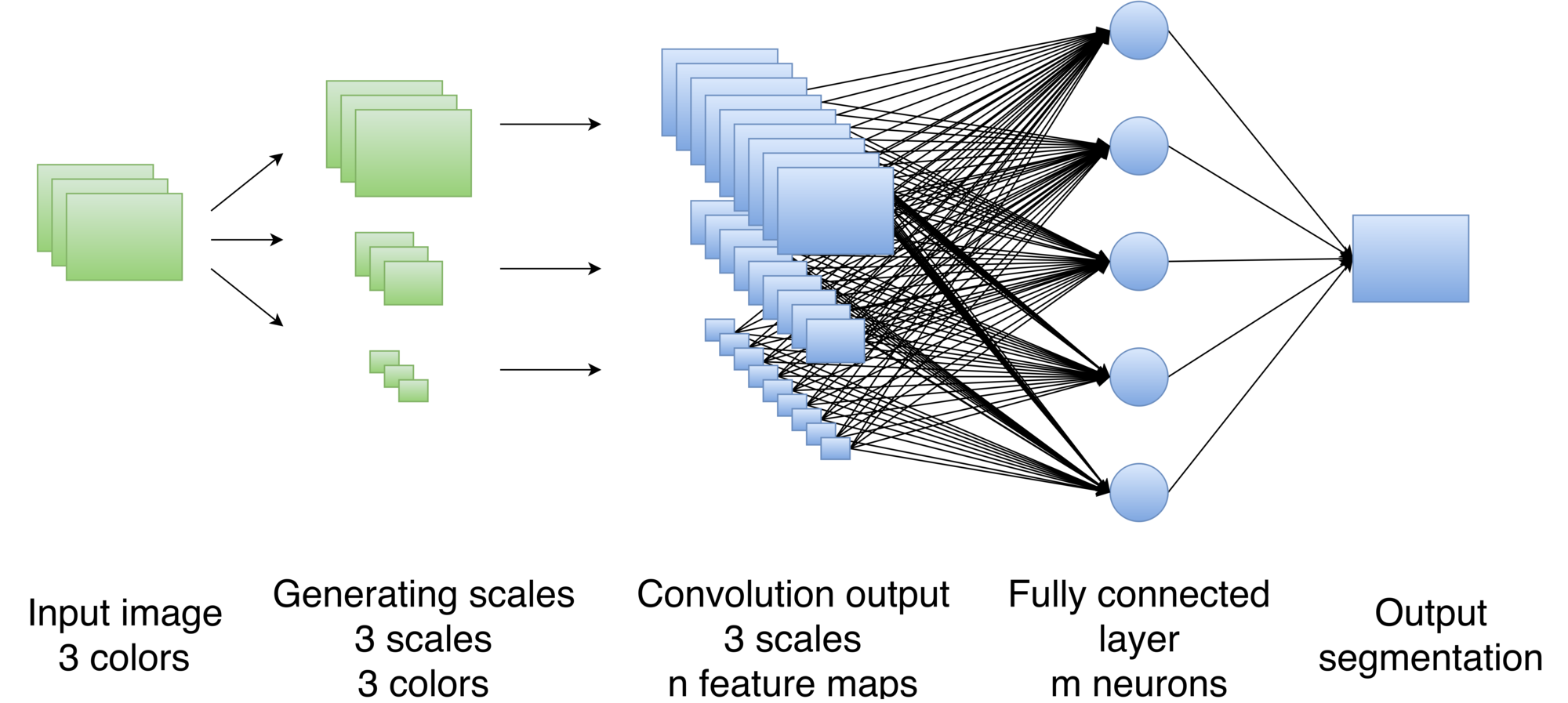}
  \end{center}
  \caption{The architecture for the two components of our network. We  extract
    features with convolutional layers without using pooling layers  and map them
    directly to the output segmentation with a fully connected layer. For clarity,
    we show  only one convolutional layer,  and both the number  of feature maps
    $n$  and  the  number of  neurons  in  the  fully  connected layer  $m$  are
    underrepresented.}
  \label{classifier}
\end{figure}

As  shown in  Figure~\ref{classifier}, our  initial  network was  made of  three
chains of three convolution layers each.  The first chain is directly applied to
the input image, the second one to the input image after downscaling by a factor
two, and the last one to the input image after downscaling by a factor four.  We
do not use pooling  layers here, which allows us to extract  the fine details of
the hand masks.

The outputs of  these three chains are concatenated together  and given as input
to a fully connected logistic regression layer, which outputs for each pixel its
probability of lying on a hand.

\subsection{Splitting the Network in Two}
\label{sssec:splitting}

The network described  above turned out to be too  computationally intensive for
real-time.   To speed  it  up, we  developed  an approach  that  is inspired  by
Auto-Context~\cite{Tu09}.   Auto-Context is  a  segmentation method  in which  a
segmenter is iterated,  taking as input not  only the image to  segment but also
the segmentation result  of the previous iteration.   The fundamental difference
between  our  approach  and  the  original  Auto-Context  is  that  the  initial
segmentation is performed on a downscaled version of the input image.

\begin{figure}
  \begin{center}
    \includegraphics[width=\columnwidth]{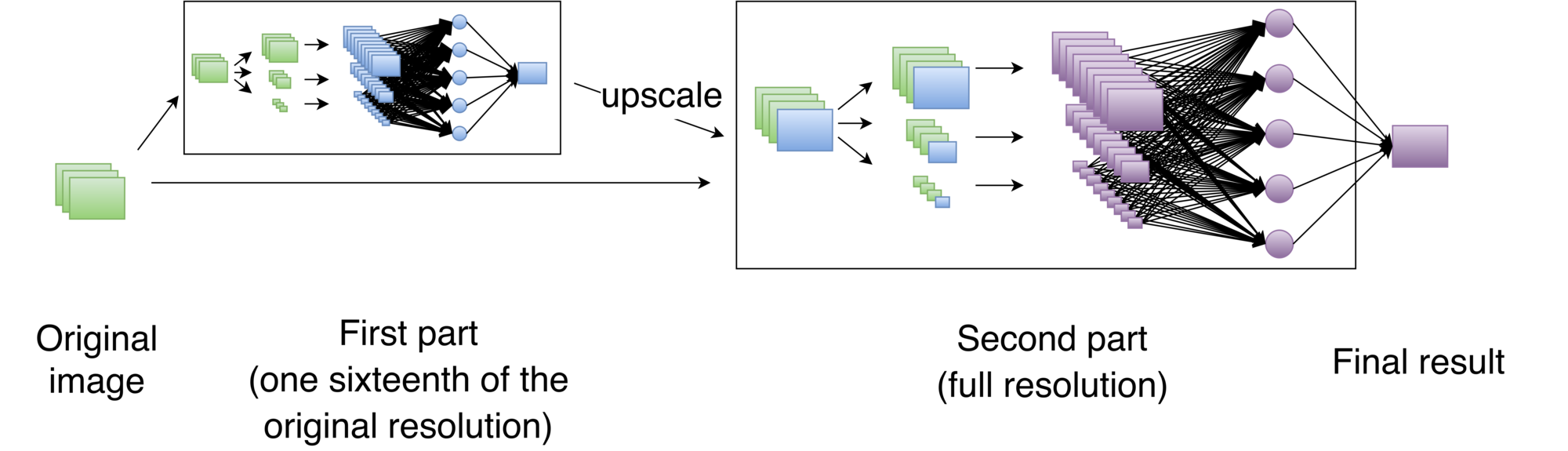}
  \end{center}
  \caption{Two-part       network      architecture.        As      in
    Figure~\ref{classifier},  only one  convolutional layer  is shown,
    and the  number of feature maps  and the number of  neurons on the
    fully connected  layer are underrepresented  in both parts  of the
    classifier to  make the  representation more  understandable.  The
    second part  of the network  receives as  input the output  of the
    first  part after  upscaling but  also the  original image,  which
    helps segmenting fine details.}
  \label{classifier2}
\end{figure}

As seen in Figure~\ref{classifier2}, the first part performs the segmentation on
the original image after downscaling by a factor 16, and outputs a result of the
same resolution.  Its output along with the original image is then used as input
to the  second part of  the new  network, which is  a simplified version  of the
initial network  to produce  the final,  full-resolution segmentation.   The two
parts of  the network have very  similar structures. The difference  is that the
second part  takes as input the  original, full resolution input  image together
with the  output of  the first  part after upscaling.  The first  output already
provides a  first estimate of  the position of the  hands; the second  part uses
this information in  combination with the original image  to effectively segment
the image. An example of the feature maps computed by the first part can be seen
in Figure~\ref{maps}.

The advantage of this split is two-fold:  The first part runs on a small version
of the original image, and we can considerably reduce the number of feature maps
and use smaller filters in the second part without loosing accuracy.

\subsection{Meta-Parameters Selection}

There is currently no good way to determine the optimal filter sizes and numbers
of feature maps, so  these have to be guessed or determined  by trial and error.
For  this reason  we trained  networks with  the same  structure, but  different
parameters  and compared  the accuracy  and running  time.  We  first identified
parameters that  produced the  best results while  ignoring processing  time and
then simplified  the model  to reduce  processing time  while retaining  as much
accuracy as possible.

Our input images have a  resolution of 752$\times$480, scaled to 188$\times$120,
94$\times$60, and 47$\times$30 and input to  the first part of the network.  The
first  two layers  of each  chain output  32 feature  maps and  the third  layer
outputs 16  feature maps.  We used  filters of size 3$\times$3,  5$\times$5, and
7$\times$7 pixels  for the successive  layers.  For  the second part,  the first
layer outputs  8 feature maps,  the second layer 4  feature maps, and  the third
layer outputs  the final probability  map.  We  used 3$\times$3 filters  for all
layers.

We used the leaky rectified linear unit as activation function~\cite{leaky_ref}.
We  minimize  a  boosted  cross-entropy  objective  function~\cite{cost}.   This
function weights the samples with lower probabilities more. We used $\alpha = 2$
as  proposed  in  the  original   paper.   We  used  RMSprop~\cite{rmsprop}  for
optimization.

To avoid overfitting and to make the classifier more robust, we augmented the
training set using very simple geometric transformations: We used scaling by a
random factor between 0.9 and 1.1, rotating for up to 10 degrees, introducing
shear for up to 5 degrees, and translating by up to 20 pixels.

\begin{figure}
  \begin{center}
    \begin{tabular}{cc}
      \includegraphics[width=38mm]{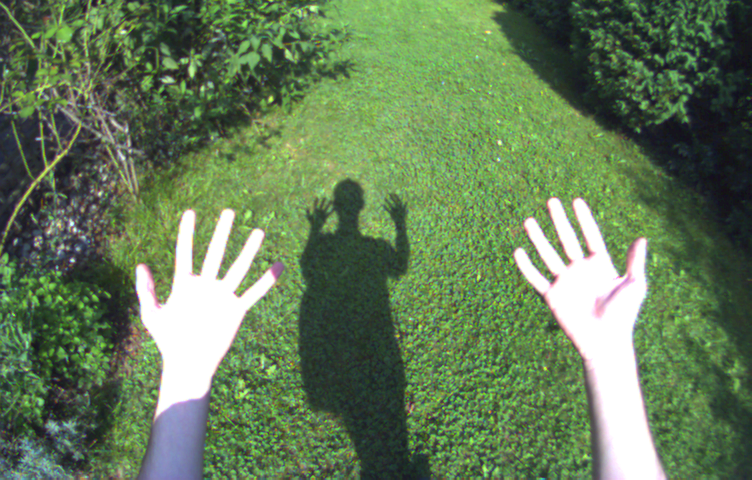}
      &
      \includegraphics[width=38mm]{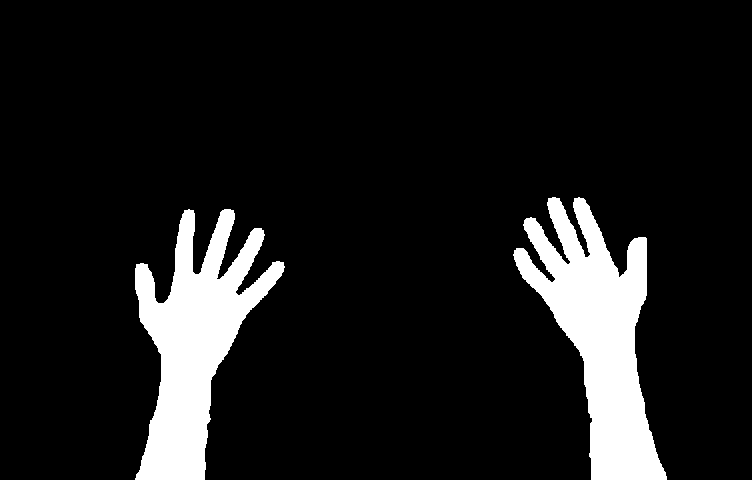}
    \end{tabular}
    \begin{tabular}{cccc}
      \includegraphics[width=17mm]{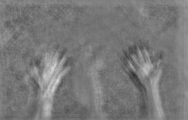} &
      \includegraphics[width=17mm]{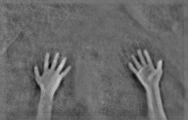} &
      \includegraphics[width=17mm]{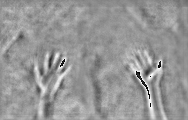} &
      \includegraphics[width=17mm]{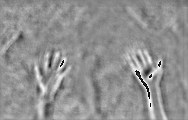} \\ 
      
      \includegraphics[width=17mm]{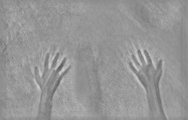} & 
      \includegraphics[width=17mm]{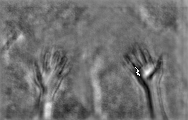} &
      \includegraphics[width=17mm]{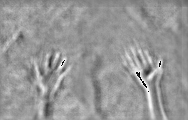} &
      \includegraphics[width=17mm]{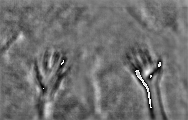} 
    \end{tabular}
  \end{center}
  \caption{Original input  image, its  ground truth  segmentation, and
    some of the  resulting feature maps computed by the  first part of
    the network.}
  \label{maps}
\end{figure}

\begin{figure}
  \begin{center}
    \includegraphics[width=\columnwidth]{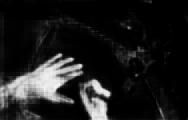}
  \end{center}
  \caption{Example  of output  of the  first part  from the  final architecture.
    Shades of grey represent the probabilities  of the hand class over the pixel
    locations.}
  \label{reduced_resolution}
  \label{maps}
\end{figure}

\section{Results}

In this section, we describe the dataset we built for training and testing our
approach, and present and discuss the results of its evaluation.

\subsection{Dataset}

\begin{figure}
  \begin{center}
    \begin{tabular}{ccc}
      \includegraphics[width=25mm]{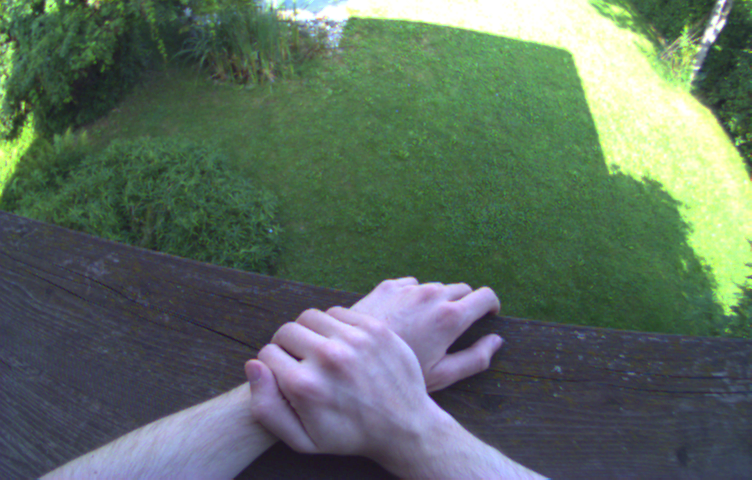} &
      \includegraphics[width=25mm]{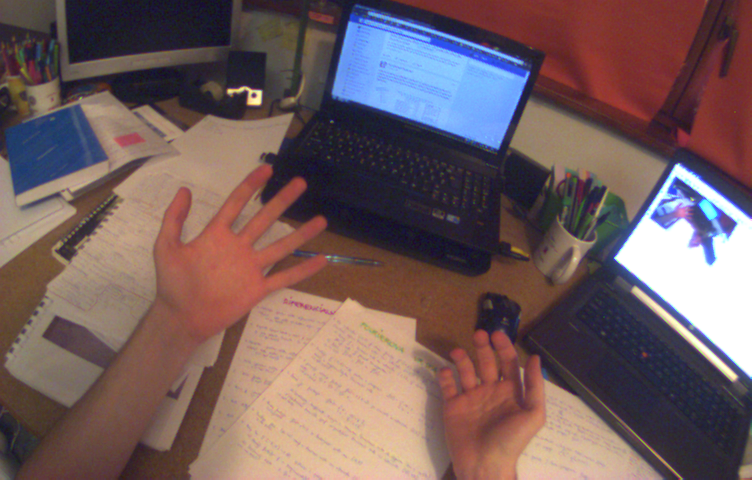} &
      \includegraphics[width=25mm]{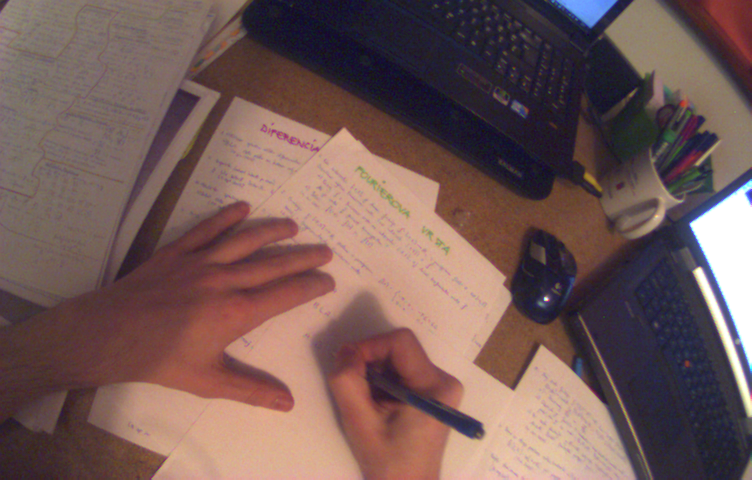} \\ 
      
      \includegraphics[width=25mm]{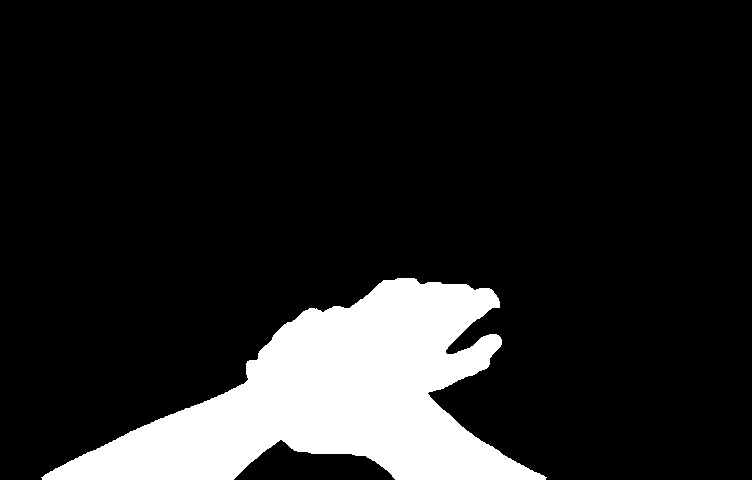} &
      \includegraphics[width=25mm]{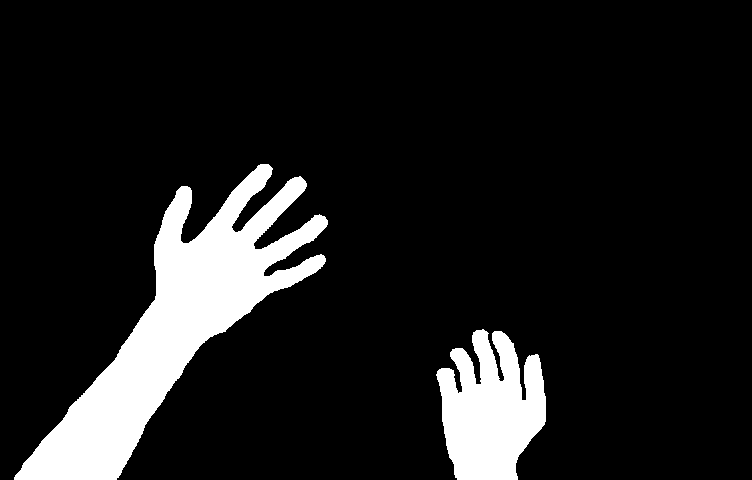} &
      \includegraphics[width=25mm]{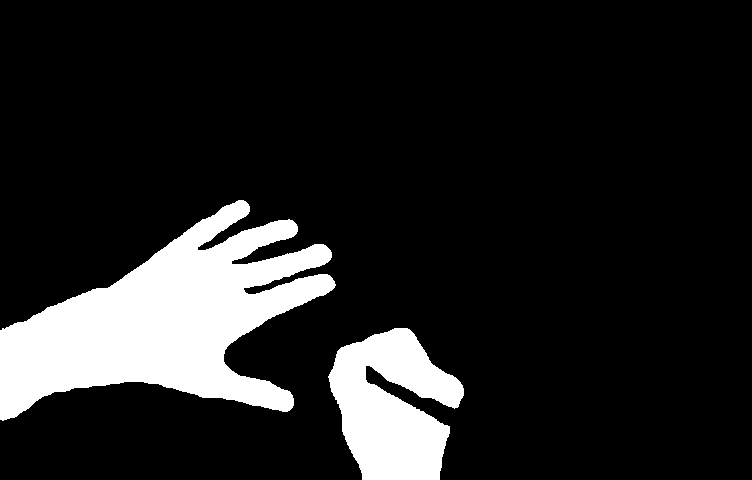} \\ 
      
      \includegraphics[width=25mm]{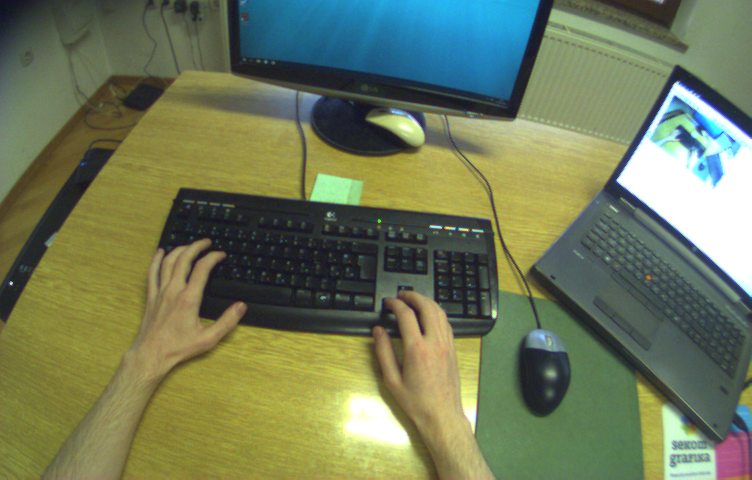} & 
      \includegraphics[width=25mm]{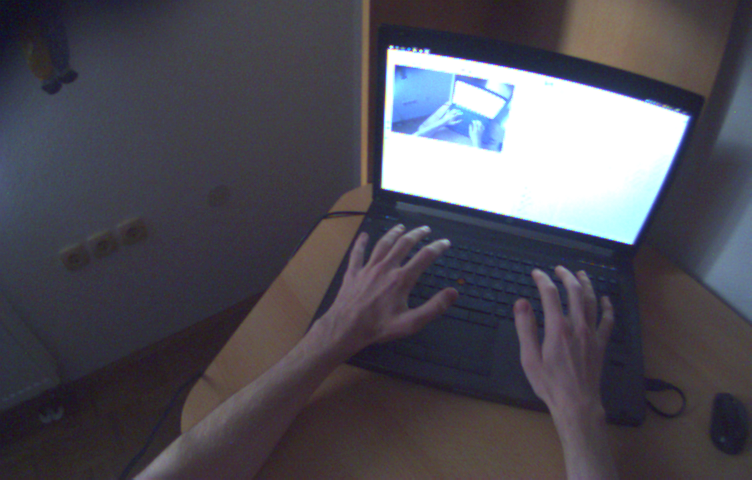} &
      \includegraphics[width=25mm]{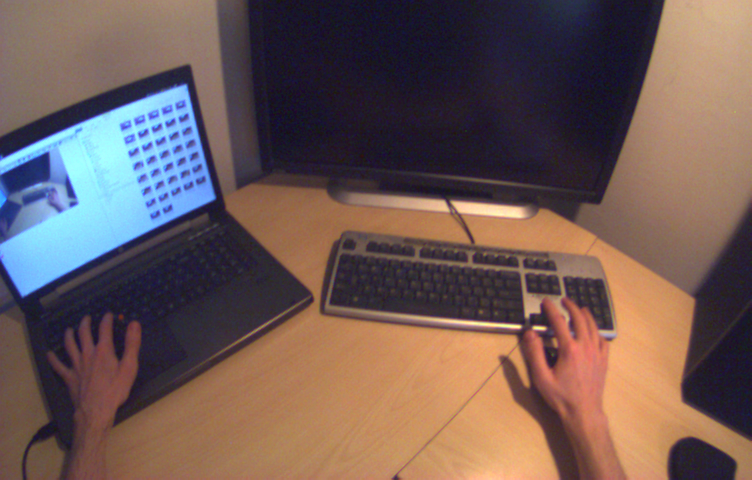} \\

      \includegraphics[width=25mm]{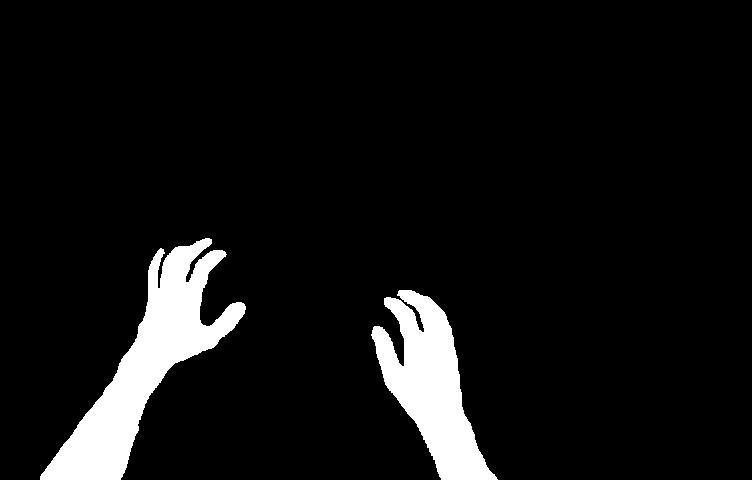} & 
      \includegraphics[width=25mm]{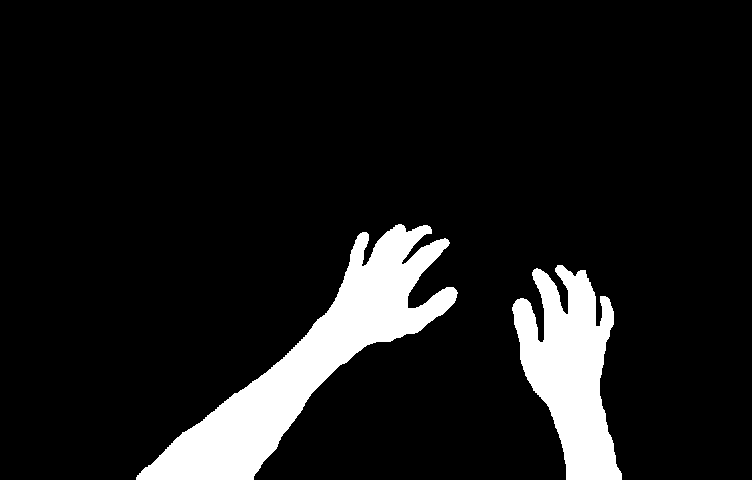} &
      \includegraphics[width=25mm]{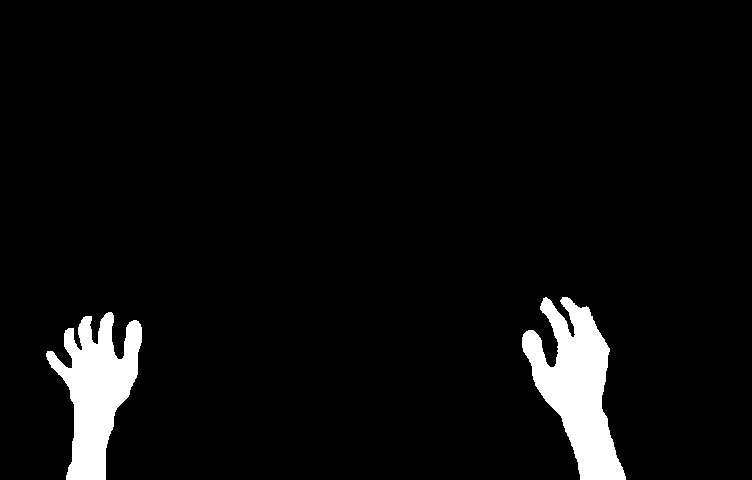}
    \end{tabular}
  \end{center}
  \caption{Some of the images from our dataset and their ground truth segmentations.}
  \label{images_ground}
\end{figure}

We  built a  dataset  of samples  made  of  pairs of  images  and their  correct
segmentation   performed   manually.   Figure~\ref{images_ground}   shows   some
examples.

We focused  on egocentric images, i.e.   the hands are seen  from a first-person
perspective.  Several subjects  acquired these images using  a wide-angle camera
mounted on  their heads,  near the eyes.   The camera was  set to  take periodic
images of whatever  was in the field of  view at that time. In  total 348 images
were taken. 90\% of the images were used for training, and the rest was used for
testing.

191 of  those images  were taken  in an  office at  6 different  locations under
different  lighting conditions.   The remaining  157  images were  taken in  and
around  a residential  building, while  performing everyday  tasks like  walking
around, opening  doors etc.   The images  were taken  with an  IDS MT9V032C12STC
sensor with resolution of 752$\times$480 pixels.




\begin{figure}
  \begin{center}
    \includegraphics[width=\columnwidth]{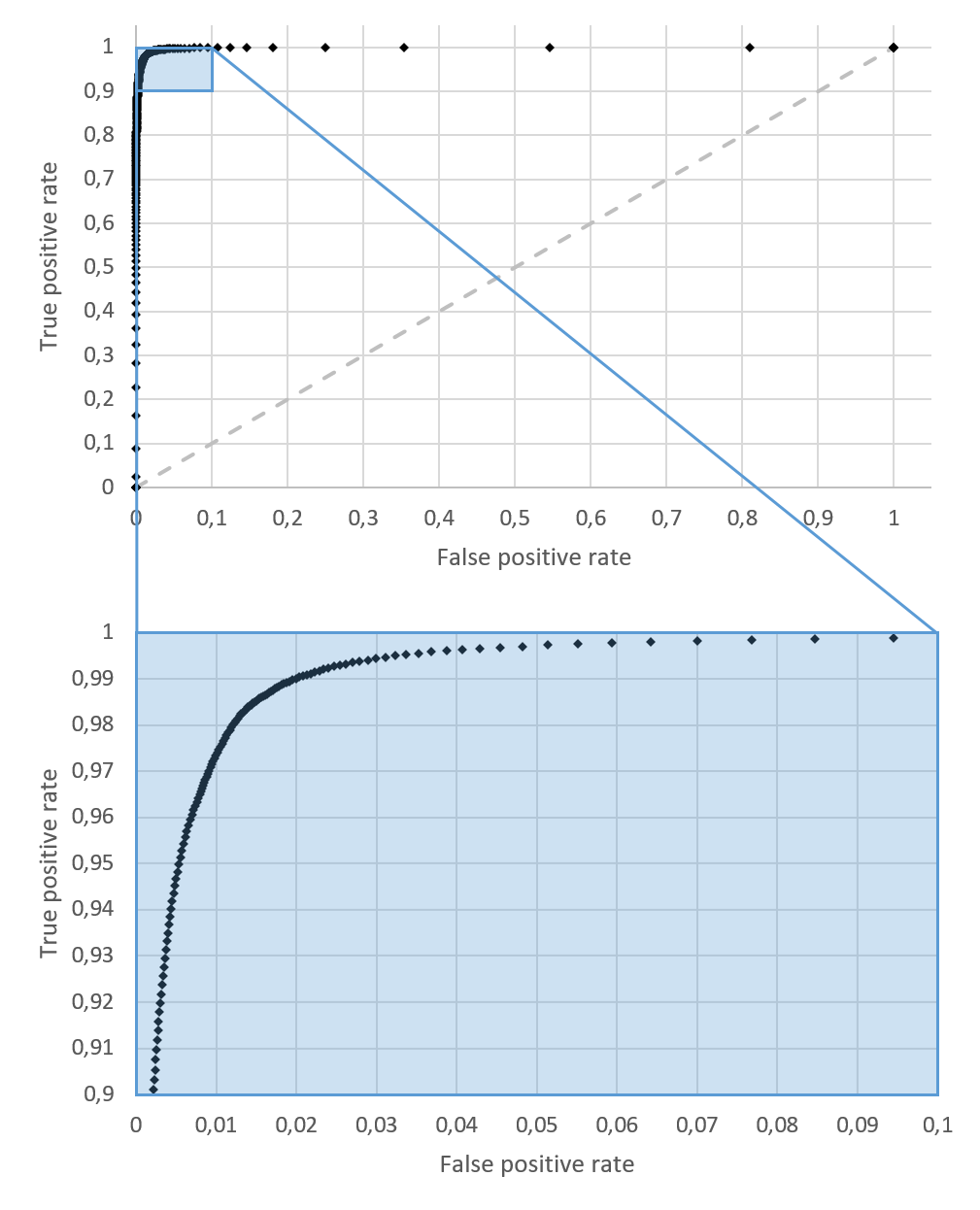}
  \end{center}
  \caption{ROC  curve obtained  with our  method on  our challenging  dataset of
    egocentric images.  The  figure also shows a magnification  of the top-left
    corner.}
  \label{roc_curve}
\end{figure}

\begin{figure}
  \begin{center}
    \includegraphics[width=\columnwidth]{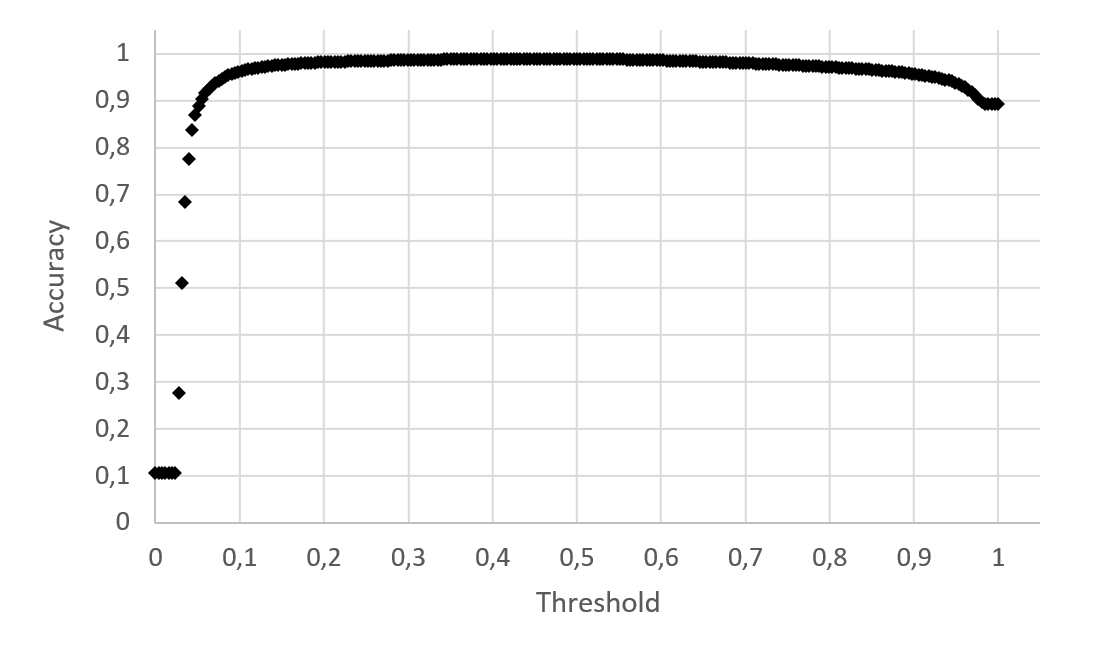}
  \end{center}
  \caption{Accuracy  of  the  classifier  depending  on  probability  threshold.
    Best accuracy can be obtained for a large range of thresholding values.}
  \label{accuracy}
\end{figure}


\subsection{Evaluation}

\begin{figure}
  \begin{center}
    \begin{tabular}{ccc}
      \includegraphics[width=25mm]{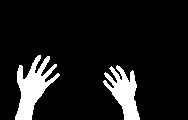} &
      \includegraphics[width=25mm]{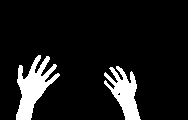} &
      \includegraphics[width=25mm]{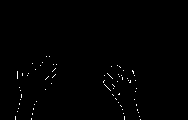} \\ 
      
      \includegraphics[width=25mm]{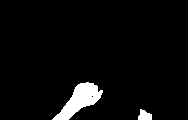} &
      \includegraphics[width=25mm]{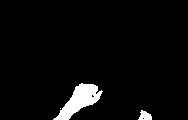} &
      \includegraphics[width=25mm]{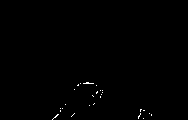} \\
      
      \includegraphics[width=25mm]{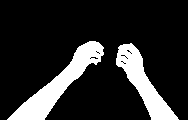} &
      \includegraphics[width=25mm]{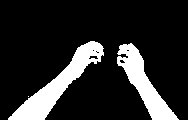} &
      \includegraphics[width=25mm]{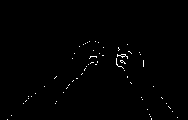} \\ 
      
      \includegraphics[width=25mm]{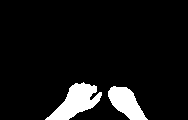} & 
      \includegraphics[width=25mm]{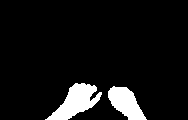} &
      \includegraphics[width=25mm]{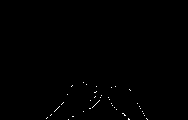} \\
      (a) & (b) & (c)\\      
    \end{tabular}
  \end{center}
  \caption{Comparison  between  the  ground   truth  segmentations and the
    predicted ones.  (a) Ground truth, (b) prediction, (c) differences. Errors
    are typically very small, and 1-pixel thin.}
  \label{diff}
\end{figure}

\begin{figure*}
  \begin{center}
    \begin{tabular}{cccc}
      \includegraphics[width=38mm]{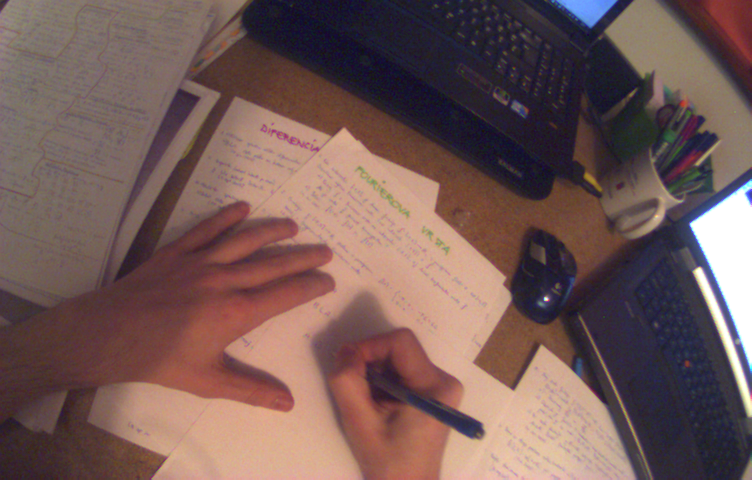} &
      \includegraphics[width=38mm]{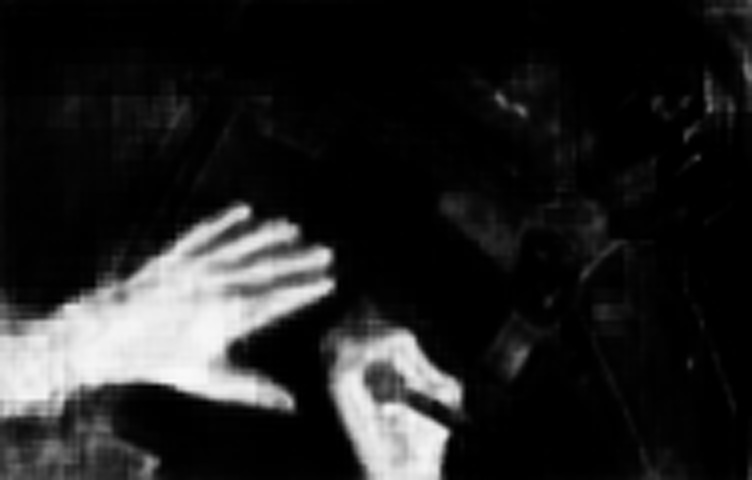} &
      \includegraphics[width=38mm]{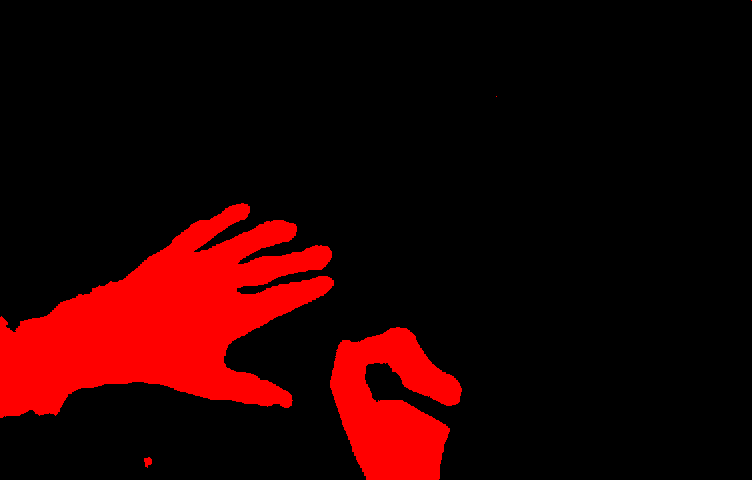} &
      \includegraphics[width=38mm]{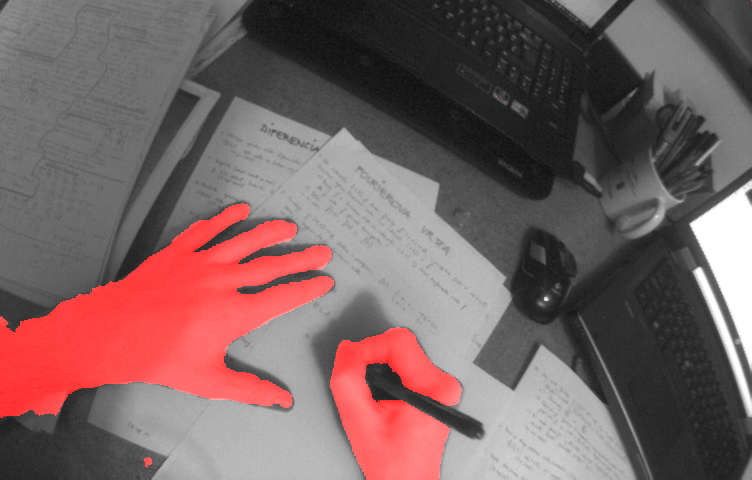} \\ 
      
      \includegraphics[width=38mm]{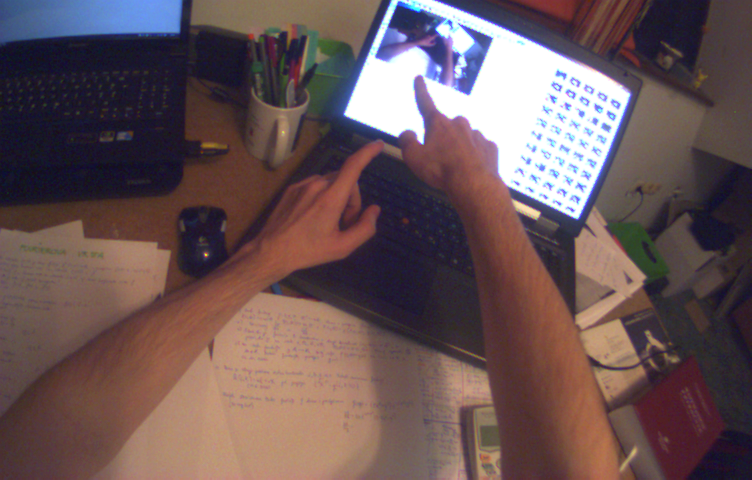} & 
      \includegraphics[width=38mm]{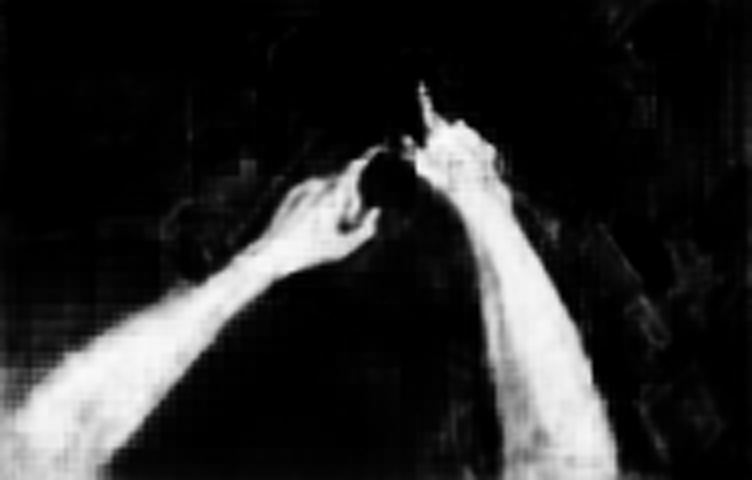} &
      \includegraphics[width=38mm]{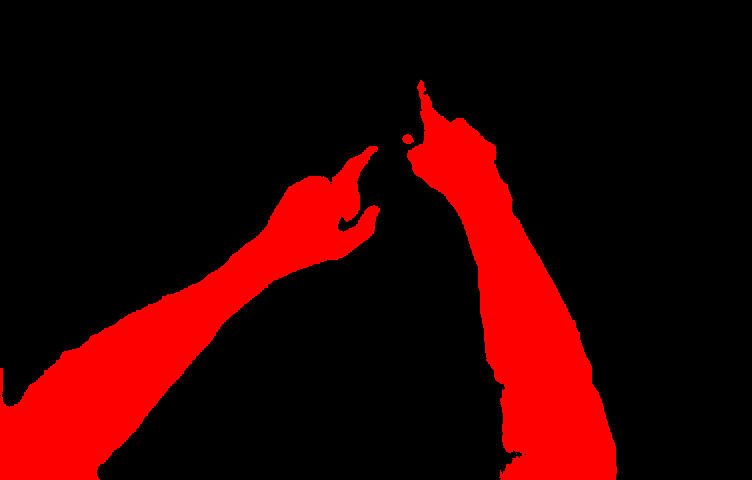} &
      \includegraphics[width=38mm]{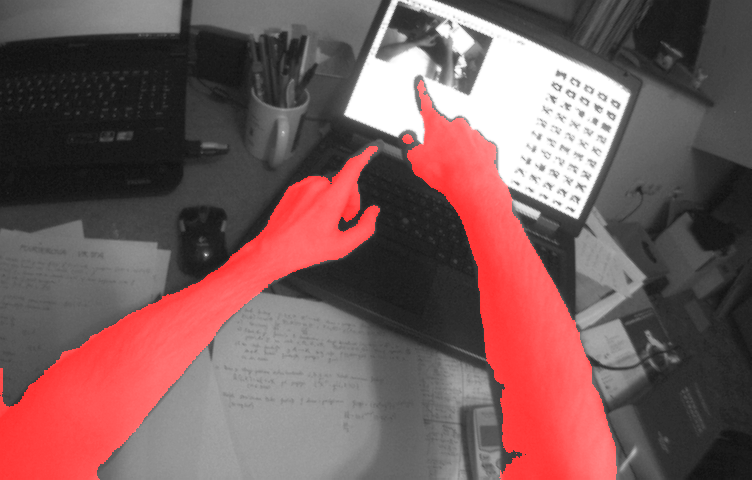}  \\ 
      
      \includegraphics[width=38mm]{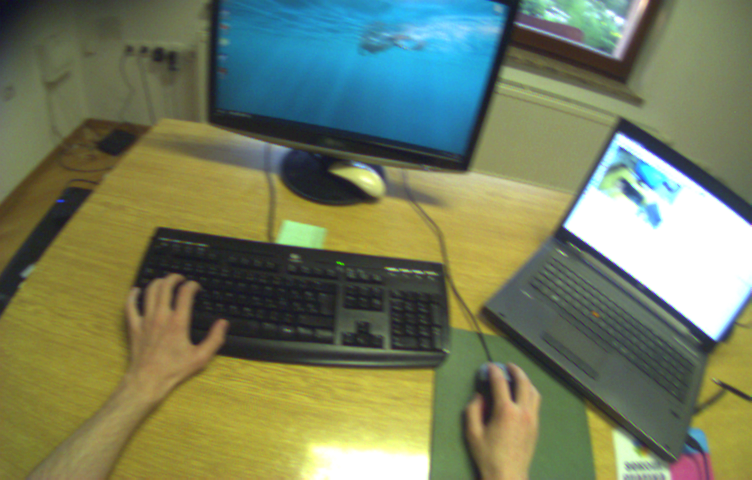} & 
      \includegraphics[width=38mm]{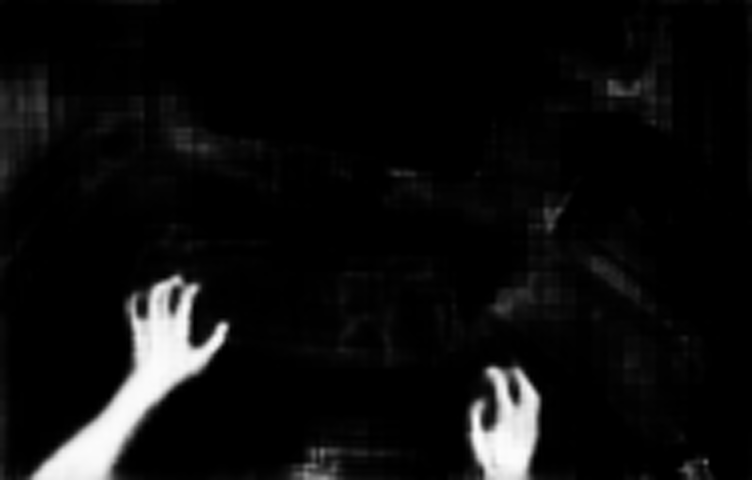} &
      \includegraphics[width=38mm]{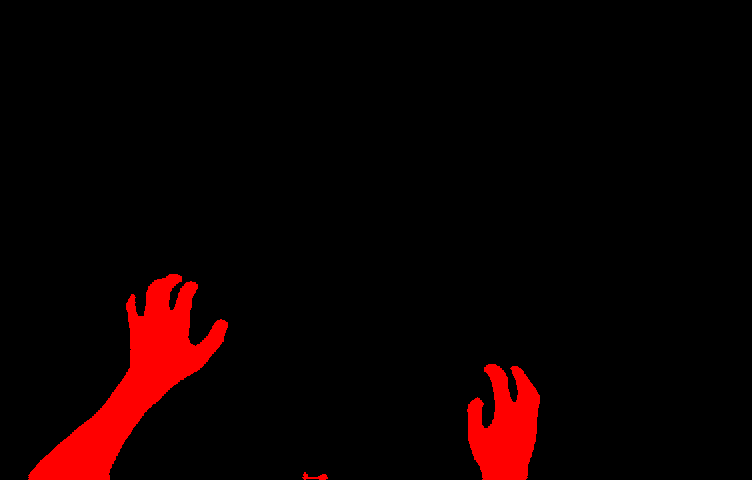} &
      \includegraphics[width=38mm]{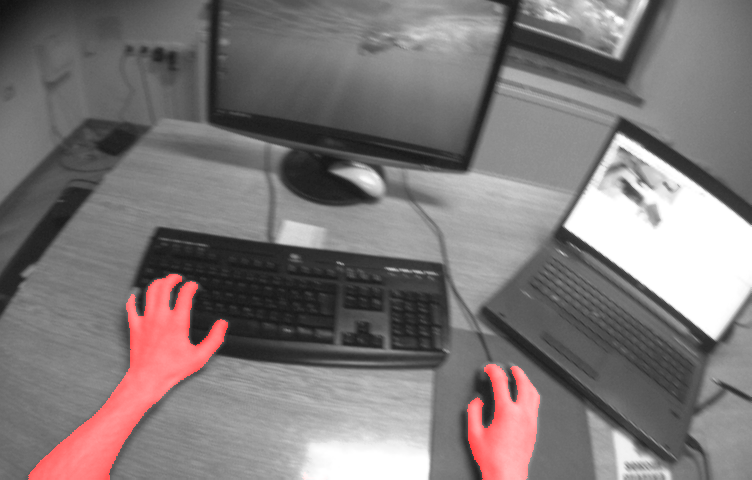}  \\ 
      
      \includegraphics[width=38mm]{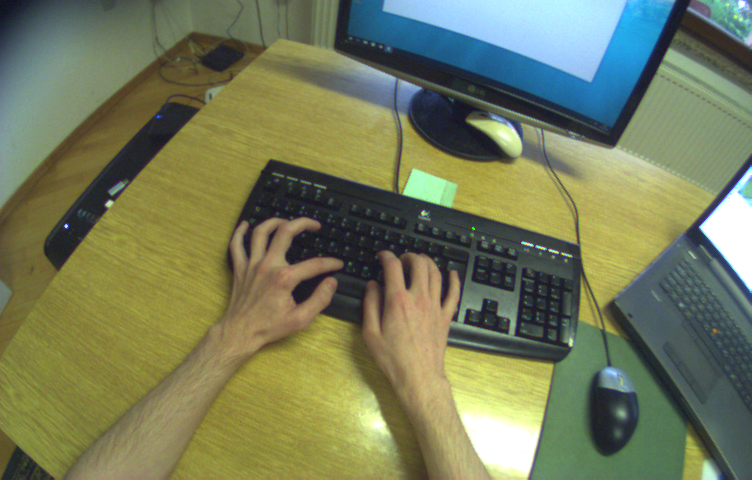} & 
      \includegraphics[width=38mm]{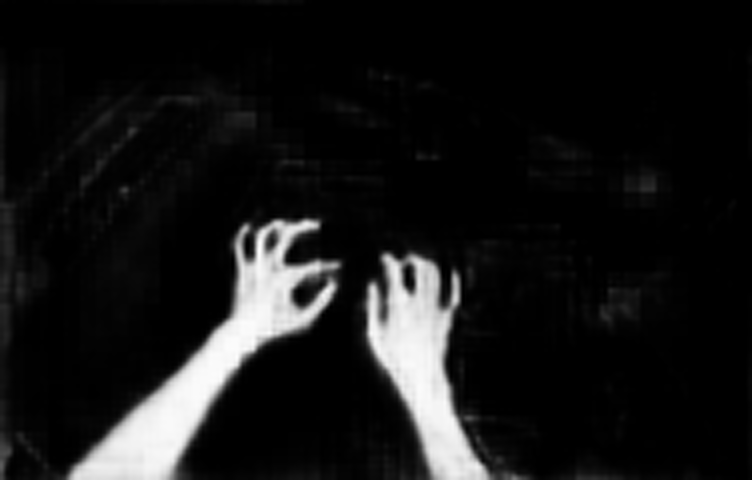} &
      \includegraphics[width=38mm]{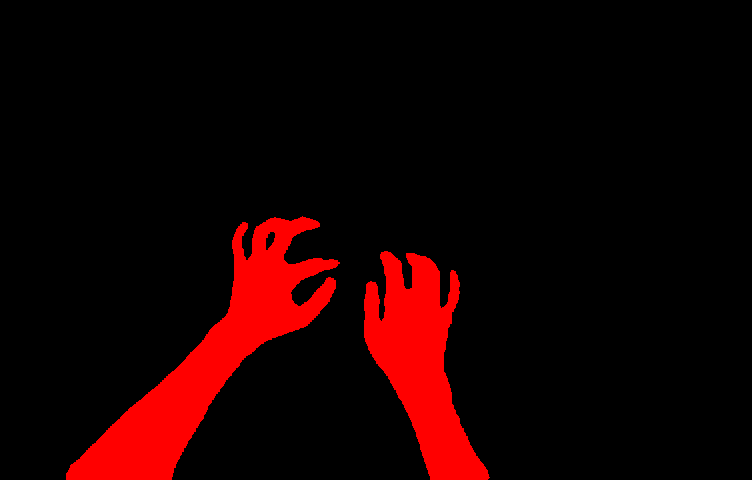} &
      \includegraphics[width=38mm]{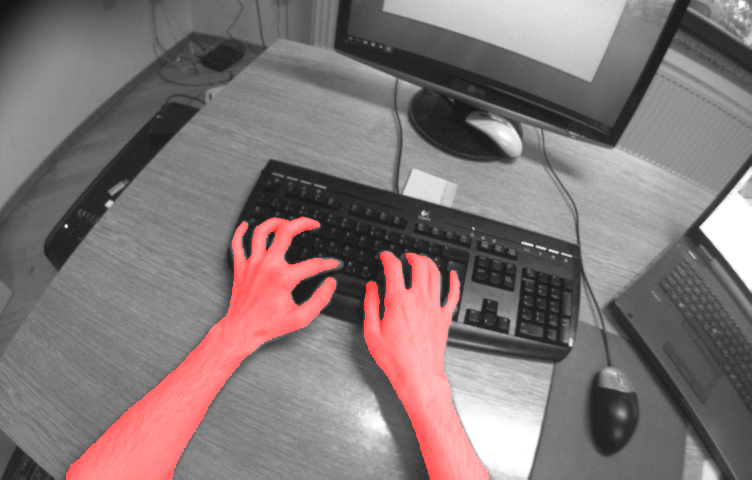}  \\ 
      
      \includegraphics[width=38mm]{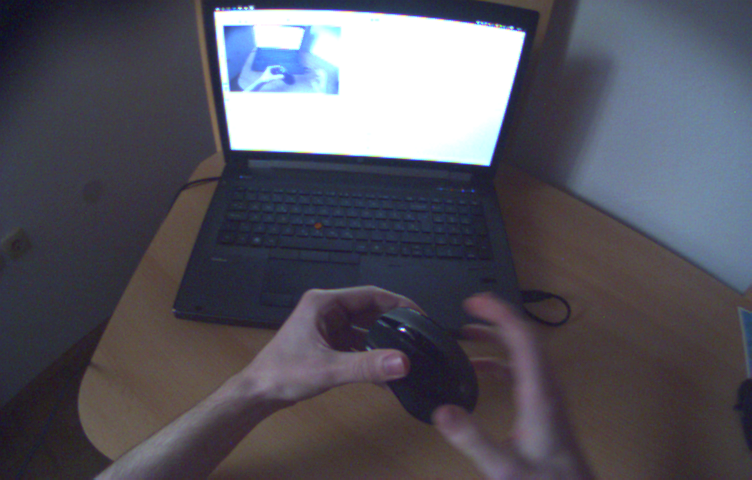} & 
      \includegraphics[width=38mm]{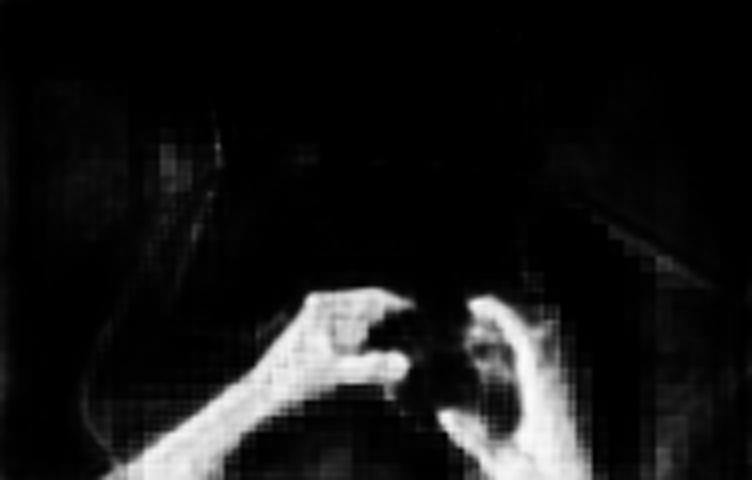} &
      \includegraphics[width=38mm]{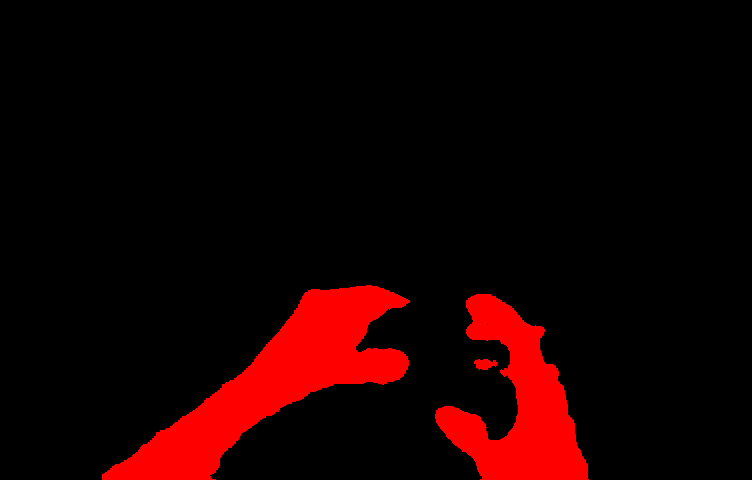} &
      \includegraphics[width=38mm]{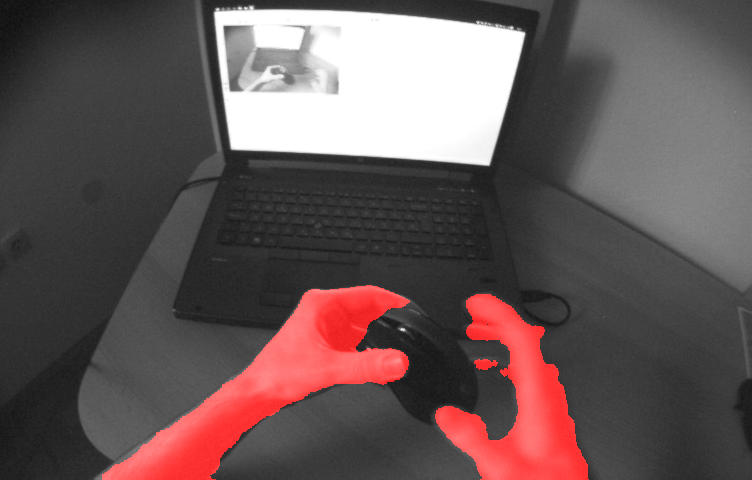}
      \\

      (a) & (b) & (c) & (d)\\
    \end{tabular}
  \end{center}
  \caption{Images  and their  segmentations.  (a) Original  image; (b)  Upscaled
    segmentation  predicted  by  the  first  part  of  our  network;  (c)  Final
    segmentation; (d)  composition of the  final segmentation into  the original
    image.}
  \label{result}
\end{figure*}

Figure~\ref{roc_curve}  shows  the ROC  curve  for  our  method applied  to  our
egocentric  dataset.  When  applying a  threshold of  50\% to  the probabilities
estimated by our method, we achieve a 99.3\% accuracy on our test set, where the
accuracy is defined as the percentage of pixels that are correctly classified.
Figure~\ref{accuracy} shows that this accuracy can be obtained with thresholds
from a large range of values, which shows the robustness of the method.
Qualitative results can be seen in Figures~\ref{diff} and \ref{result}.




\subsection{Meta-parameter Fine Selection}

In  total,  we trained  98  networks  for  the reduced  resolution  segmentation
estimation and 95 networks for the full resolution final classifier.

After meta-parameter fine  selection for the first part of  the network, we were
able to achieve the  accuracy of 98.3\% at 16 milliseconds  per image, where the
first  layer had  32 feature  maps and  3$\times$3 filter,  the second  layer 32
feature maps and 5$\times$5 filters, and the third layer had 16 feature maps and
7$\times$7 filters.

With further meta-parameters fine selection for  the second part of the network,
we obtained a network reaching 99.3\%  for a processing time of 39 milliseconds.
The first  layer (of the  second part)  had 8 feature  maps, the second  layer 4
feature maps,  and the  third layer  1 feature map.  All layers  used 3$\times$3
filters.


During  this fine  selection  process  we noticed  that  the  best results  were
achieved when  the number of  filters was higher  for earlier layers  and filter
sizes  were  bigger  at  later  layers.   The  reduced  resolution  segmentation
estimation already provided very good results,  but it still produced some false
positives.  Because  of the  lower resolution  the edges were  not as  smooth as
desired.  The full resolution final classifier was in most cases able to improve
both the false positives and produce smoother edges.

\subsection{Evaluation of the Different Aspects of the Method}

\subsubsection{Convolution on Full Resolution Without Pooling and Upscaling}

To verify  that splitting the classifier  into two parts performs  better than a
more standard  classifier, we  trained a classifier  to perform  segmentation on
full  resolution  images  without   first  calculating  the  reduced  resolution
segmentation estimation.

We used the same structure as the  second part of our classifier and modified it
to only  use the  original image. To  compensate the absence  of input  from the
first part, we  tried using more feature  maps. The best trade-off  we found was
using 16 feature maps and filter size  of 5$\times$5 pixels on each of the three
layers---instead of 3$\times$3  and 8, 4, and 1 feature  maps. Because of memory
size limit on the used GPU, we were not able to train a more complex classifier,
which may produce  better results.  Nevertheless, processing time  per image was
185 milliseconds  with accuracy  of 94.0\%, significantly  worse than  the proposed
architecture.


\subsubsection{Upscale Without the Original Image}

To verify that  the second part of the classifier  benefited from re-introducing
the original image compared to only having results of the first part, we trained
a classifier like the one suggested in  this work, but this time we provided the
second part  of the  classifier with  only results  of the  first part.  In this
experiment processing time was 36.7 milliseconds, compared to 39.2 milliseconds in the
suggested classifier and the accuracy fell from 99.3\% to 98.6\%. Processing
was therefore faster, but the second part of the classifier was not able to
improve the accuracy much further. The second part of the classifier was able to
correct some false positives  from the first part, but   unable to improve accuracy
along the edges between foreground and background.

\subsection{Comparison to a Color-based Classification}

As discussed in  the introduction, segmentation based on skin  color is prone to
fail as other parts of the image  can have similar colors.  To give a comparison
we applied  the method described  in \cite{skin_color_segmentation} to  our test
set and  obtained an  accuracy of  81\%, which is  significantly worse  than any
other approach we tried.


\section{Conclusion}

Occlusions are crucial  for understanding the position of  objects. In Augmented
Realist applications, their exact detection  and correct rendering contribute to
the feeling  that an object is  a part of the  world around the user.  We showed
that starting with a low resolution  processing of the image helps capturing the
context of the  image, and using the  input image a second  time helps capturing
the fine details of the foreground.

\bibliographystyle{abbrv}
\bibliography{string,vision,graphics,refs}

\end{document}